# Neural reconstruction of 3D ocean wave hydrodynamics from camera sensing

Jiabin Liu[#,*,1,2], Zihao Zhou[#,1,2], Jialei Yan[1,2], Anxin Guo[1,2], Alvise Benetazzo[*,3], Hui Li[*,1,2]

[1] Key Lab of Smart Prevention and Mitigation of Civil Engineering Disasters of the Ministry of Industry and Information Technology, Harbin Institute of Technology, Harbin, 150090, China

[2] Key Lab of Structures Dynamic Behavior and Control of the Ministry of Education, Harbin Institute of Technology, Harbin, 150090, China

[3] Institute of Marine Sciences, National Research Council, Venice, Italy

**Abstract**

Precise three-dimensional (3D) reconstruction of wave free surfaces and associated velocity fields is essential for developing a comprehensive understanding of ocean physics. To address the high computational cost of dense visual reconstruction in long-term ocean wave observation tasks and the challenges introduced by persistent visual occlusions, we propose an wave free surface visual reconstruction neural network, which is designed as an attention-augmented pyramid architecture tailored to the multi-scale and temporally continuous characteristics of wave motions. Using physics-based constraints, we perform time-resolved reconstruction of nonlinear 3D velocity fields from the evolving free-surface boundary. Experiments under real-sea conditions demonstrate millimetre-level wave elevation prediction in the central region, dominant-frequency errors below 0.01 Hz, precise estimation of high-frequency spectral power laws, and high-fidelity 3D reconstruction of nonlinear velocity fields, while enabling dense reconstruction of two million points in only 1.35 s. Built on a stereo-vision dataset, the model outperforms conventional visual reconstruction approaches and maintains strong generalization in occluded conditions, owing to its global multi-scale attention and its learned encoding of wave propagation dynamics.

## 1. Introduction

Wave dynamics serve as a primary driver in nearshore physical ocean systems, where their complex spatiotemporal evolution governs energy fluxes[1], coastal material transport[2], and boundary layer processes[3]. Amid increasingly frequent extreme marine events driven by global climate change[4], the accurate and effective measurement of wave hydrodynamic parameters has emerged as a forefront challenge in advancing observational capabilities and predictive

---

# Jiabin Liu and Zihao Zhou contributed equally to this work

* Corresponding author.

*E-mail address*: liujiabin@hit.edu.cn, alvise.benetazzo@ve.ismar.cnr.it, lihui@hit.edu.cn

accuracy in ocean physics[5, 6]. Traditional measurement techniques predominantly depend on contact-based sensors, such as wave staffs and moored wave buoys[7], which are capable of delivering stable and long-duration time series of surface elevation data[8]. However, the inherently limited spatial resolution of these point-based instruments impedes the comprehensive characterization of the continuous, evolution of wave fields[9]. This severely constrains the ability to model and forecast complex hydrodynamic processes at regional scales[10].

With the deployment of large arrays of low-cost cameras, recent advances in vision-based remote sensing methods offer promising alternatives[11]. In particular, computer vision approaches enable non-intrusive, high-resolution capture of both spatial wave free surface structures and temporal wave dynamics[12]. Among these, stereo vision (SV) has been widely applied in both laboratory and field environments for wave reconstruction through stereo matching and triangulation[13, 14]. Subsequent developments have optimized the reconstruction algorithms using the OpenCV library, incorporating self-calibration based on the scale-invariant feature transform (SIFT)[15, 16] and employing the semi-global block matching (SGBM)[17] algorithm to enhance reconstruction accuracy. A notable example is the wave acquisition stereo system (WASS), which utilizes a fixed stereo camera configuration to enable large-scale, high-density wave free surface reconstruction[18]. Further integration of stereo vision with digital image processing allows for frame-by-frame acquisition of wave imagery using synchronized cameras, enabling the construction of a spatial-temporal volume for sequential wave free surface reconstruction[19]. Experimental results confirm that the reconstructed wave free surfaces exhibit strong agreement with classical wave theory in terms of spatial-temporal distribution and dynamic characteristics, thereby validating the reliability and physical consistency of this vision-based measurement approach[20].

Vision-based techniques for dynamic surface reconstruction encounter significant challenges. As the resolution and coverage of imaging systems increase, the number of pixels and disparity hypotheses grows rapidly, imposing substantial demands on memory and processing time and limiting the feasibility of real-time or continuous observation, particularly in field applications with constrained computational resources. For megapixel-level dense reconstruction, recovering the free surface at a single time instant typically requires several minutes of computation. This substantial computational burden has become a critical bottleneck for large-scale deployment and sustained ocean wave monitoring. Pixel correspondence is also highly sensitive to imaging perturbations such as occlusion, strong illumination, and overexposure. These issues are especially pronounced in oceanic environments, where varying meteorological and lighting conditions can drastically alter the visual scene. For example, storm events may introduce severe visual occlusions due to waves and spray, while clear-sky conditions may generate regions of overexposed pixels. Conventional vision methods (e.g., BM[21], GC[22], SGBM[17]), which rely on local feature matching, are intrinsically limited in handling such challenges, as they often fail to exploit the inherent spatial and temporal continuity of dynamic surfaces, making it difficult to achieve a complete and temporally coherent reconstruction of continuously evolving wave fields. Various approaches—including tracer particles[23], projected structured light[24], laser imaging[25], or calibration grids[26]—have been explored to enhance surface measurability; however, these often involve complex deployment procedures and limited adaptability in open-sea conditions, restricting their practicality for long-term, autonomous wave monitoring.

Recent advances in deep learning have brought new opportunities for 3D reconstruction[27, 28]. Depth estimation from a fixed viewpoint typically yields only relative depth, limiting its ability to reconstruct metrically accurate 3D structures[29, 30]. While methods like NeRF[31] and Gaussian splatting[30] can recover absolute scale, they require dynamic views or multi-angle observations, making them unsuitable for fixed marine monitoring setups. In contrast, stereo vision offers a more practical solution for accurate 3D reconstruction under such constraints. Stereo matching networks such as RAFT-Stereo[32], UniMatch[33], Selective-Stereo[34] and YOLO-based[35] models have shown notable improvements in stereo estimation accuracy and computational efficiency. Current neural network models for vision-based reconstruction of wave free surfaces face persistent limitations. In particular, they are generally constrained to recovering the mean water level, while failing to resolve fine-scale surface details. This limitation is largely due to the disparity of scales: observation distances typically range from 10–100 m, whereas wave motions are on the order of meters, with smaller fluctuations down to the centimeter level. As a result, high-frequency, small-amplitude variations in both space and time are often lost in the reconstruction process. Furthermore, variations in environmental conditions, including complex ocean illumination, wave-breaking occlusions, and atmospheric effects such as rain or fog, impose greater demands on the generalization of these models.

In this study, we introduce an ocean wave hydrodynamic visual sensing neural network (WHVS-Net) that addresses the fundamental challenges of weak textures, occlusions, and temporal inconsistencies inherent in vision-based reconstruction. WHVS-Net employs an attention-augmented pyramidal architecture to capture the multi-scale and temporally continuous characteristics of free surface dynamics. Wave features are matched in the latent space through the fusion of global image information and camera calibration parameters, while multi-scale attention and temporal consistency modeling enhance robustness in feature extraction and matching. Building on the reconstructed free surface, the framework enables nonlinear 3D velocity field recovery, thereby advancing high-fidelity characterization of wave hydrodynamics. Validation under real-sea conditions demonstrates the accuracy and generalization of WHVS-Net, establishing a data-driven paradigm for intelligent and scalable visual sensing of ocean waves.

## 2. Results

*2.1 Implementation details*

The WASS project[18] is a long-term in situ wave observation project that provides raw images from the Black Sea coast (44.39°N, 33.98°E)[36, 37] and is used in study to train our network under realistic conditions. The geographical location map is shown in Fig. 1a. This region is influenced by multiple wind systems and constrained by complex topography, exhibiting rich nonlinear wave behavior and strong directional scattering characteristics. The acquisition system was installed on a coastal platform and comprised a fixed-baseline stereo camera array oriented toward a nearshore monitoring zone[36]. The configuration provides wide-angle spatial coverage with synchronized perspectives, while a hardware synchronization mechanism ensures the stable acquisition of continuous stereo image pairs. The raw dataset includes time-synchronized grayscale stereo images, along with the intrinsic and extrinsic calibration parameters of the cameras[36]. Each image has a resolution of 2456 × 2058 pixels.

Unlike traditional stereo datasets such as KITTI[38], ScanNet[39], and SceneFlow[40], which focus on rigid or semi-rigid scenes, WHVS-Net training requires data that capture highly dynamic, non-rigid, and continuously deforming surfaces. Developed under the WASS[36] project using the SGBM algorithm, the high-resolution dataset of wave images encompasses diverse sea states and observational conditions, providing a standardized benchmark for the validation of wave free surface reconstruction. In the stereo vision reconstruction workflow (Fig. 1b), disparity prediction through a deep neural network enables the recovery of point clouds in the camera coordinate system. Most existing models focus purely on image-based reconstruction (e.g., disparity or depth)[41, 42] and thus lack the physical interpretability required for extracting hydrodynamic parameters like wave spectra. To achieve a more physically meaningful representation, a geometric self-supervised plane fitting module is employed to estimate the free-surface reference plane, defined by the mean water level ($Z = 0$). The resulting rotation matrix and translation vector are then used to transform the reconstructed points from the camera coordinate system into a global coordinate system, facilitating the extraction of the 3D wave free surface (Fig. 1c). Following the transformation and alignment (Fig. 1d), the reconstructed wave elevation $\eta$ is projected onto the left image view within the defined reconstruction region.

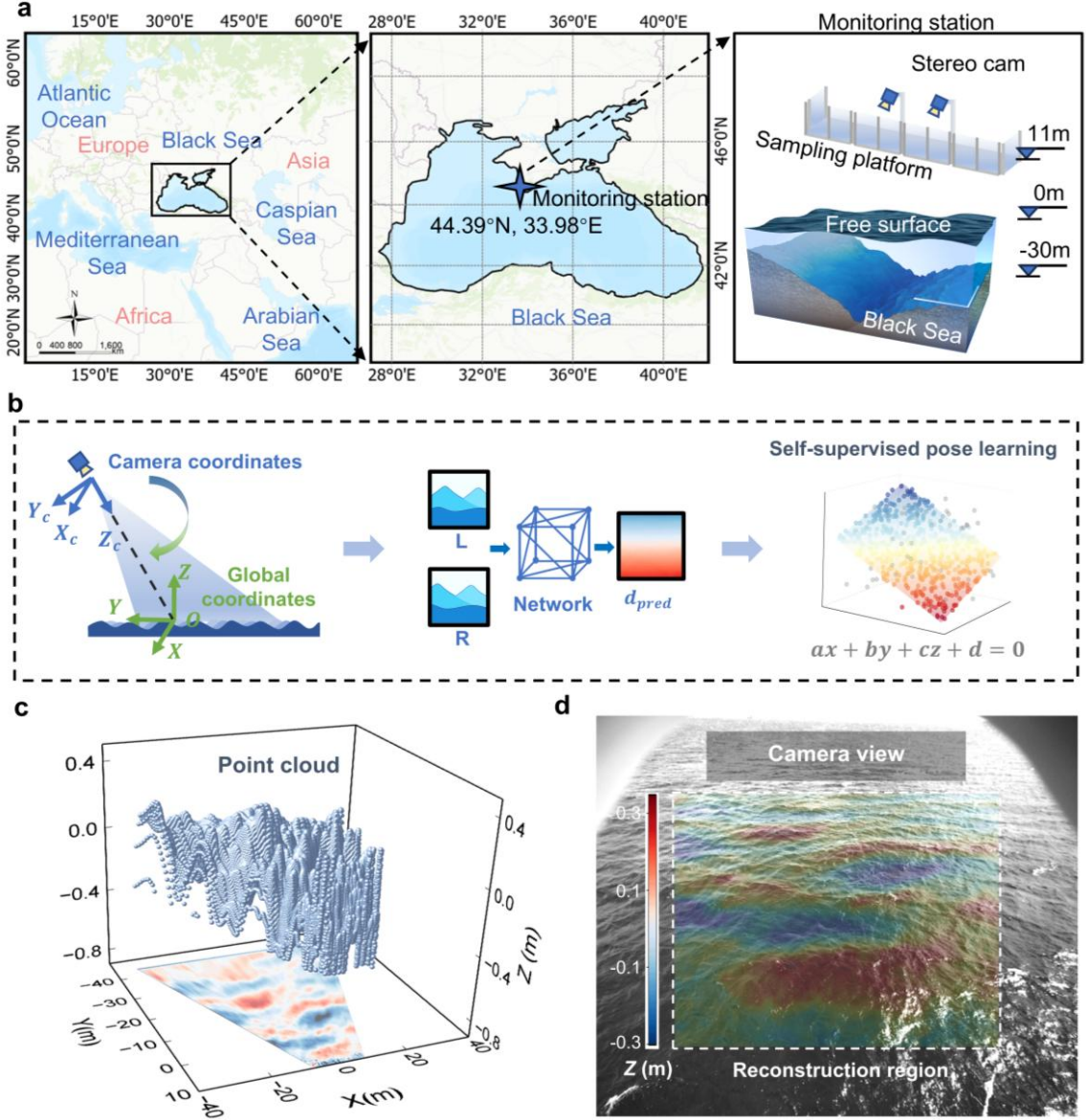

**Fig. 1** Stereo vision-based acquisition and reconstruction of wave free surface. **a** Geographical context of the sampling site, showing the Black Sea and its surrounding seas, with a magnified view indicating the observation location at 44.39°N, 33.98°E. **b** Workflow of the wave free surface reconstruction, illustrating the transformation from camera to global coordinates, disparity prediction through a deep network, and plane fitting via self-supervised pose learning. **c** Point clouds for dense 3D surface reconstruction. **d** Camera-view example of a reconstructed ocean wave free surface, with the defined reconstruction region highlighted. The reconstruction image region is defined as $u \in [412, 2044], v \in [389, 1669]$, where $u$ and $v$ are the horizontal and vertical pixel coordinates, respectively. This yields a final resolution of 1632 × 1280 pixels, corresponding to approximately 2 million reconstructed points.

The dataset comprises continuous stereo sequences that capture diverse wave free surface patterns and their spatiotemporal evolution, exhibiting highly dynamic characteristics. During training, the left and right images $I_L(u, v)$ and $I_R(u, v)$ are jointly fed into the network to generate a predicted disparity map $d_{\text{pred}}(u, v)$, which is compared against the ground-truth $d_{\text{GT}}(u, v)$ to compute the loss and supervise the optimization of network parameters. The training data are collected under varying wave conditions[36], with each condition containing approximately 10,000–40,000 images. These temporal segments exhibit significant changes in both illumination and wave parameters. The data are divided into training, validation, and test sets in a ratio of 10:1:1. The wave scenarios comprising the independent test set, which was held out for final evaluation, are listed in Table 1. The test set is categorized into two groups, A and B, based on variations in camera pose. Each group comprises data from three distinct time periods, labeled A1, A2, A3 and B1, B2, B3, respectively.

Table 1 Hydrodynamic conditions of the test set

| Dataset ID | $H_s$ (mm) | $f_p$ (Hz) | $T_p$ (s) | $B$ (mm) |
|---|---|---|---|---|
| A1 | 450 | 0.35 | 2.94 | 2030 |
| A2 | 360 | 0.38 | 2.63 | 2030 |
| A3 | 550 | 0.27 | 3.61 | 2030 |
| B1 | 650 | 0.19 | 5.27 | 1872 |
| B2 | 410 | 0.24 | 4.24 | 1872 |
| B3 | 660 | 0.23 | 4.44 | 1872 |

**Note**: $H_s$ denotes the significant wave height. $f_p$ and $T_p$ represent the peak frequency and wave period, respectively. $B$ is the baseline length.

*2.2 Architecture for WHVS-Net*

Inspired by the multi-scale matching strategy of pyramidal optical flow methods, we reformulate stereo matching as constrained pyramidal flow estimation[43]. Within this unified framework, the model inherits the inherent advantages of multi-scale processing—capturing large displacements at coarse resolutions and refining local details at finer scales[44]. As illustrated in Fig. 2, the proposed WHVS-Net integrates three major components to enable stereo-based reconstruction of wave sea surfaces. First, the network employs multi-scale and single-scale encoders to extract hierarchical representations from the input image frames. Through residual blocks[45], these encoders generate the left and right feature maps ($F_L$, $F_R$) along with reference feature maps ($F_s'$). Notably, the reference features are derived exclusively from the normalized left image, ensuring structural independence from the stereo feature maps and providing stable guidance for disparity estimation.

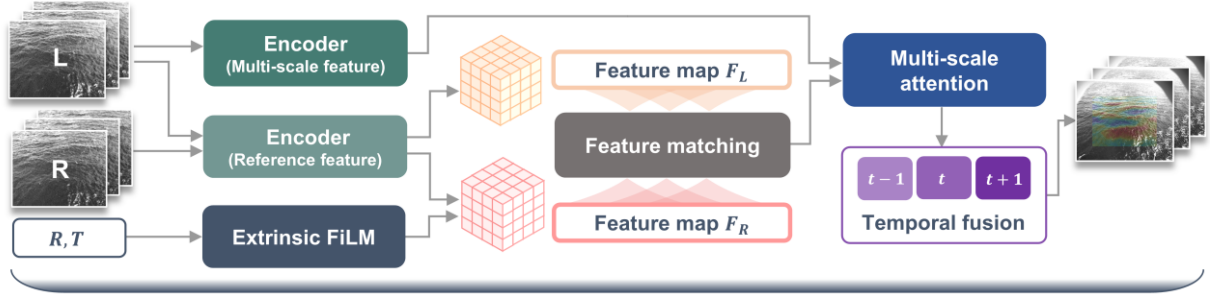

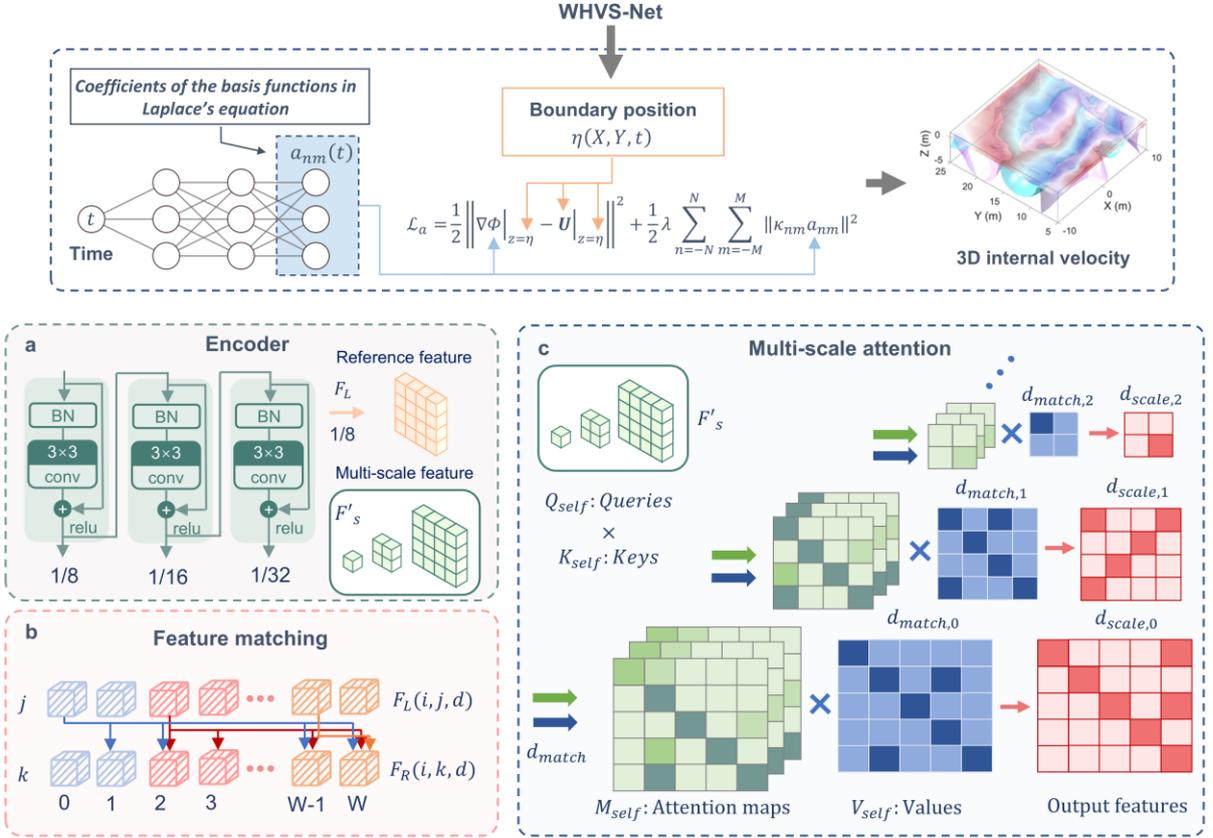

**Fig. 2** Architecture of WHVS-Net and internal 3D internal velocity reconstruction. **a** Feature encoder that extract hierarchical features from input wave images, generating feature maps ($F_L$, $F_R$) and reference feature maps ($F'_s$), through residual blocks. **b** One-dimensional feature matching module computes spatial correspondences between left and right features to guide disparity estimation ($d_{match,s}$). **c** Multi-scale attention mechanism is applied to refine disparity predictions across different spatial resolutions. Attention maps are computed using queries and keys projected from the reference feature maps ($F'_s$), while the values are constructed from disparity cues ($d_{match,s}$). This process yields disparity estimates ($d_{scale,s}$) at different scales, which are subsequently integrated in the decoder to produce the final disparity outputs. The details of the 3D internal velocity reconstruction are presented in Subsection 2.5.

An extrinsic feature-wise linear modulation (FiLM) modulation mechanism is introduced, which encodes the camera's rotation and translation parameters into channel-wise scaling and shifting coefficients that are applied to the feature maps. This modulation enables the feature representations to adapt to varying camera poses. Based on this feature representation, a one-dimensional correlation module is applied to perform feature matching along epipolar lines. By operating directly in the latent space, this module establishes pixel-wise correspondences between the left and right feature maps, thereby producing preliminary disparity cues ($d_{match,s}$). Such latent-space matching not only ensures differentiability and computational efficiency, but also enables the network to implicitly learn geometric consistency while suppressing local ambiguities that are difficult to resolve in the image domain. These cues serve as the foundation for hierarchical disparity estimation.

To address the long-standing trade-off between capturing large displacements and preserving fine structural details, WHVS-Net incorporates a multi-scale attention mechanism operating across progressively downsampled feature maps at resolutions of 1/8, 1/16, and 1/32 [32]. Within this mechanism, queries and keys are projected from the reference feature maps ($F'_s$), while values are derived from the disparity cues ($d_{match,s}$). The attention operation refines disparity predictions across scales, yielding multi-resolution disparity maps ($d_{scale,s}$). Ocean wave is temporally smooth. Therefore, introducing a temporal fusion mechanism can effectively suppress instantaneous noise and enhance the physical consistency of the predictions.

The architecture implements a hierarchical feature fusion mechanism that enables coarse-to-fine disparity estimation, explicitly tailored to the non-rigid and multi-scale characteristics of ocean waves. This design balances the dual objectives of capturing large, dynamic displacements inherent in wave dynamic while reconstructing fine structural details of the free surface, making it more suitable for wave hydrodynamic sensing under real sea conditions. Details are provided in the *Methods*.

*2.3 Reconstruction validation*

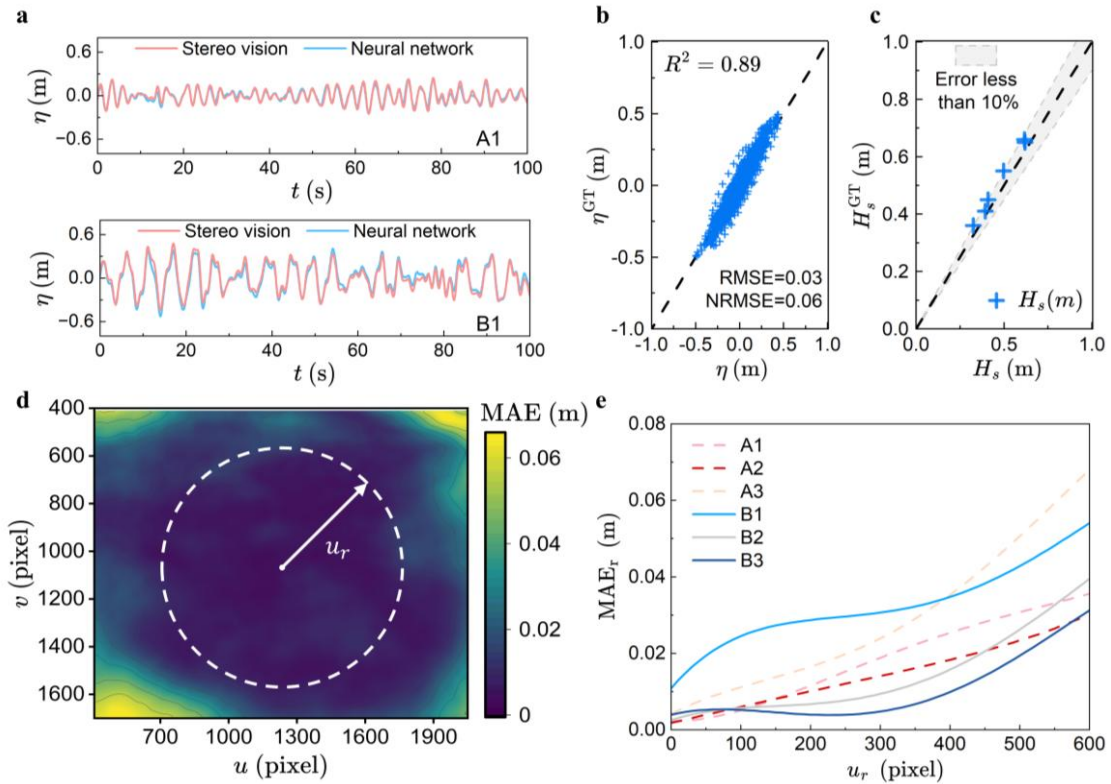

Fig. 3 Validation of wave free surface elevation reconstruction. **a** Comparison of temporal wave elevation signals reconstructed by the proposed neural network (blue) and stereo vision reference data (red) under two

different camera poses (A1 and B1). **b** Scatter plot analysis for condition A1, indicating the correlation between predicted ($\eta$) and ground-truth ($\eta^{\text{GT}}$) wave heights, with performance metrics including $R^2$, RMSE and NRMSE. **c** Significant wave height ($H_s$) error analysis across all six experimental conditions (A1–A3, B1–B3). **d** Two-dimensional distribution of mean absolute error (MAE), where warmer colors denote higher local error intensities across the image plane. **e** Radial mean absolute error $\text{MAE}_\text{r}(u_r) = \frac{1}{T}\frac{1}{N}\sum_{t=1}^{T}\sum_{i=1}^{N}|\eta(u_r, \theta_i, t) - \eta^{\text{GT}}(u_r, \theta_i, t)|$ with respect to the image center across all six conditions, showing error growth as the radius ($u_r$) increases outward from the optical center.

In Fig. 3a, the pixel center of the image is selected as the monitoring point to extract a wave elevation time series, which is compared under two different camera poses (A1 and B1). The results show that the proposed method achieves strong consistency with the stereo vision reference data in both the temporal location and amplitude of wave crests and troughs. Except for a few peaks exhibiting minor deviations, the reconstructed waveforms accurately capture the unsteady evolution of surface waves. The scatter plot analysis for condition A1 (Fig. 3b) demonstrates a strong correlation between predicted wave heights and ground-truth, with a coefficient of determination of $R^2 = 0.89$. Performance metrics including $\text{RMSE} = 0.03\ m$ and $\text{NRMSE} = 0.06\ m$ further confirm the method's high precision. Significant wave height ($H_s$) comparisons across six experimental conditions (A1-A3, B1-B3) are shown in Fig. 3c, where all estimation errors remain within 10%, indicating robust adaptability across diverse observational settings.

Further examination of the spatial distribution of reconstruction errors is provided in Fig. 3d and 3e. The error field (Fig. 3d) indicates that the mean absolute error remains consistently at the millimeter level across the majority of the domain, without evidence of systematic bias. Slightly elevated errors occur near the image boundaries, likely due to geometric distortions and reduced reliability of peripheral features; nevertheless, these deviations remain within tolerable bounds. The radial distribution in Fig. 3e exhibits a monotonic growth of error with increasing distance from the optical center. The growth rate is moderate, and the overall magnitude remains uniformly controlled, thereby confirming the robustness and stability of the proposed method across the entire field of view.

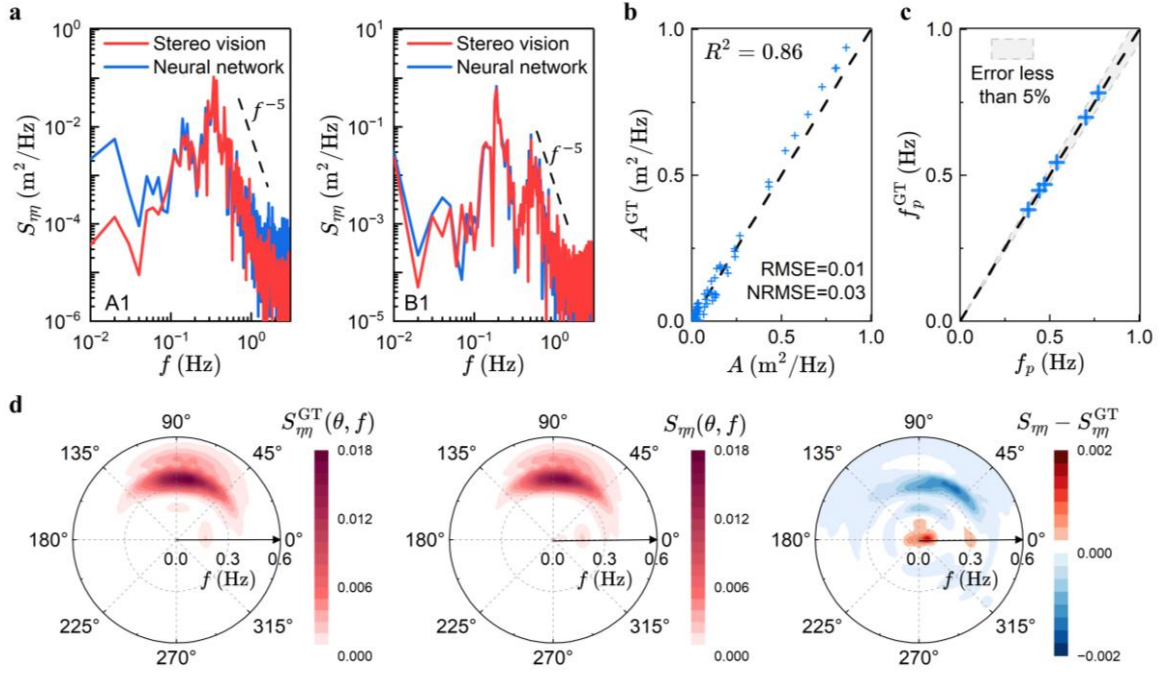

**Fig. 4** Frequency-domain analysis. **a** Power spectral density (PSD) of wave free surface elevation obtained under two different camera poses (A1 and B1). **b** Scatter plot of spectral amplitudes for condition A1 with $R^2$, RMSE, and NRMSE values. **c** Peak frequency ($f_p$) error evaluation across all six experimental conditions (A1-A3, B1-B3). **d** Two-dimensional directional frequency spectra for condition A1, including ground-truth reference ($S_{\eta\eta}^{GT}(\theta,f)$, left), predicted spectra ($S_{\eta\eta}(\theta,f)$, middle), and their residual error distribution ($S_{\eta\eta}(\theta,f) - S_{\eta\eta}^{GT}(\theta,f)$, right).

The frequency-domain validation highlights the capability of the proposed network to reproduce key spectral characteristics of oceanic waves. As shown in Fig. 4a, the predicted PSD curves closely follow the stereo vision reference across a wide frequency range, accurately capturing both the dominant peak, secondary spectral components and high-frequency components. Conventionally, the high-frequency tail of the wave spectrum is formed in equilibrium and saturation ranges [46, 47] with power laws of $f^{-4}$ and $f^{-5}$, respectively. The equilibrium range exists near the peak frequency, and the saturation range is formed at a relatively higher frequency. As shown in Fig. 4a, the visual reconstruction method effectively captures high-frequency components of the wave spectrum, which follow a power-law decay of approximately $f^{-5}$. It evident that the network model is able to learn these high-frequency fluctuations from the visually reconstructed dataset. These findings are highly promising for mitigating frequency bias in wave spectral estimation. It can be attributed to its multi-scale architecture, which facilitates the simultaneous representation of both low- and high-frequency wave components with high fidelity. Quantitative scatter plot analysis for condition A1 (Fig. 4b) demonstrates strong correlation between predicted and ground-truth spectral amplitudes, with $R^2 = 0.86$, confirming high predictive accuracy of PSD.

The accuracy of peak frequency estimation is further confirmed across all six experimental conditions (A1–A3, B1–B3), where errors remain consistently below 5% (Fig. 4c). Directional frequency spectra provide a more comprehensive assessment (Fig. 4d). The predicted spectra $S_{\eta\eta}(\theta, f)$ exhibit strong agreement with the ground-truth reference $S_{\eta\eta}^{\text{GT}}(\theta, f)$, not only reproducing the dominant wave direction (around 95°) and the associated frequency band ($f_p \approx 0.35 \text{ Hz}$), but also capturing the energy concentration patterns. Residual maps ($S_{\eta\eta}(\theta, f) - S_{\eta\eta}^{\text{GT}}(\theta, f)$) indicate that discrepancies are generally confined to low-energy peripheral regions, with maximum deviations below $2 \times 10^{-3} \text{ m}^2/\text{Hz}$. Overall, the results confirm that the proposed framework maintains high fidelity in reconstructing frequency–directional characteristics of wave fields, with errors well within acceptable limits.

*2.4 Visual occlusion*

In ocean environments, atmospheric and surface effects such as rainfall, and wave spray often cause visual occlusions that degrade image information. To assess the robustness of surface reconstruction under visual interference, we systematically evaluated two types of occlusion patterns: distribution and localization, as illustrated in Fig. 5a. Occlusions were applied simultaneously to both left and right stereo views using binary masks. Fig. 5b illustrates the non-overlapping field-of-view occlusion that arises when cropped stereo images are used as the input to the SV algorithm, leading missing disparity information near the left boundaries. Unlike the external occlusions shown in Fig. 5a, this type of data loss originates from the intrinsic baseline geometry of the stereo configuration, where the limited intersection between the two camera frustums naturally results in non-overlapping regions at the image edges.

Fig. 5c shows the variation of the mean reconstruction error with increasing occlusion ratio. For occlusion ratios below 15%, the neural network consistently outperforms conventional visual reconstruction methods, exhibiting significantly lower errors regardless of the occlusion pattern. Interestingly, under the same occlusion ratio, the network achieves better performance with localized occlusions than with distributed ones, suggesting that spatial continuity in the data facilitates accurate reconstruction.

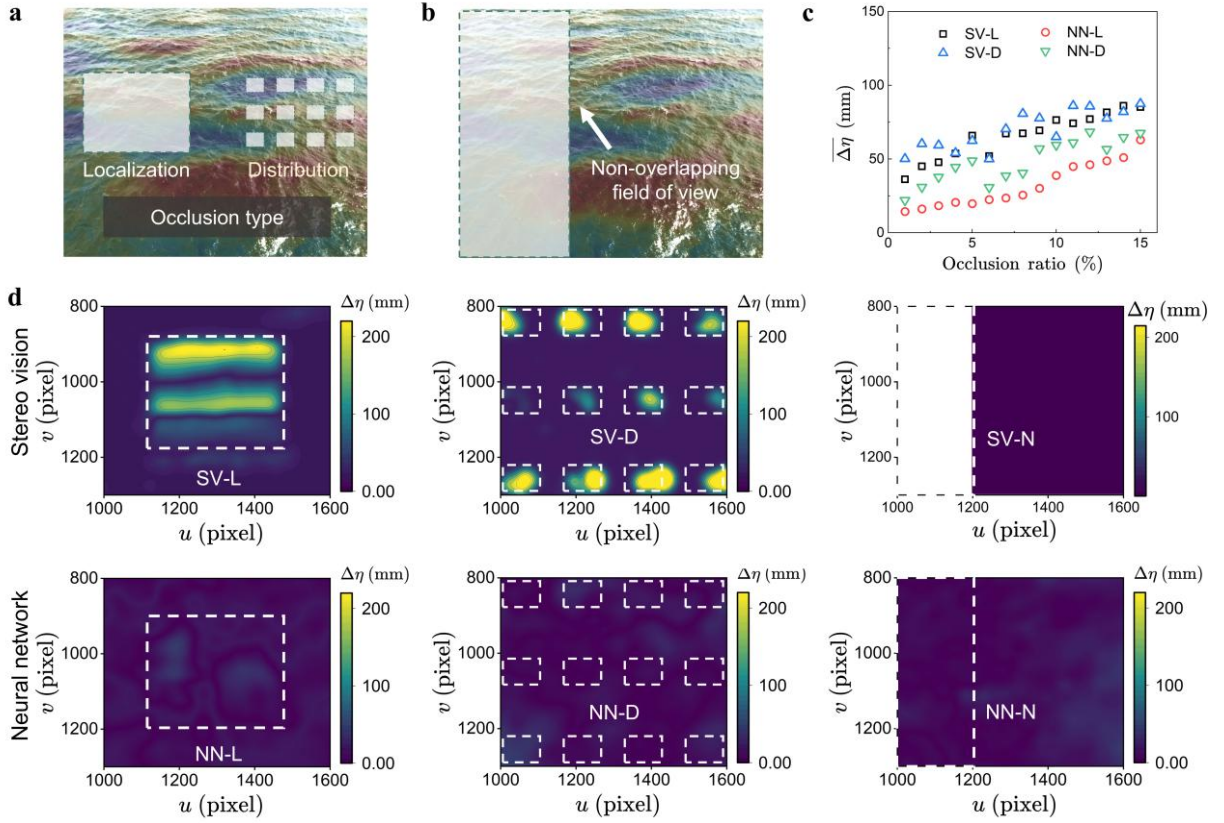

**Fig. 5 Surface reconstruction under occlusion under A1. a** Types of occlusion, including localization occlusion and distribution occlusion. **b** non-overlapping field-of-view occlusion. **c** Reconstruction error as a function of occlusion ratio, compared with ground truth. SV and NN denote stereo vision and neural network, respectively. L and D indicate localized occlusion and distributed occlusion. $\overline{\Delta \eta}$ represents the average error relative to the stereo vision reconstruction under no-occlusion conditions. **d** Error distribution of surface reconstruction using the stereo vision method and the neural network method. $\Delta \eta$ represents the per-pixel error relative to the stereo vision reconstruction under no-occlusion conditions. The error values of the NN/SV methods are reported for three occlusion scenarios applied to the same image. Note that under SV reconstruction with non-overlapping field-of-view occlusion, the blank region on the left indicates a complete loss of wave information.

Fig. 5d presents the error distribution across the reconstructed surface. The first two columns correspond to externally induced occlusions (localized and distributed). In those cases, the network achieves stable reconstruction performance despite the presence of missing disparity data. Specifically, conventional visual methods show pronounced errors in occluded regions, and even outside these regions, the reconstruction error remains higher than that of the network. This discrepancy arises because traditional visual reconstruction method, rely on pixel- or window-based correspondence estimation under brightness constancy and small disparity assumptions, making them sensitive to illumination and texture variations[17, 22, 28]. As a result, matching and propagation errors accumulate even in non-occluded regions. Errors from occluded regions further bias camera pose estimation, propagating to non-occluded areas and degrading overall reconstruction accuracy. Although such methods introduce limited global correlation via cost aggregation, they still lack the hierarchical and multi-scale feature integration inherent to neural networks. The network, in contrast, leverages both spatial and temporal correlations to wave propagate information from surrounding and temporally adjacent regions, effectively compensating for missing data in occluded areas. The last column illustrates the baseline-induced non-overlapping field-of-view occlusion, which directly causes the loss of information on the left side. This occlusion results from the inherent geometric limitation of stereo vision—regions visible to stereo camera lack valid disparity and therefore cannot be reconstructed by conventional stereo matching methods. However, the proposed network effectively mitigates this limitation by learning to maintain structural consistency and temporal continuity, even in non-overlapping regions where pixel-wise correspondence is unavailable.

Notably, the network was trained solely on unoccluded data. Its ability to generalize and reconstruct occluded surfaces demonstrates that the network can extract and integrate intrinsic global patterns, achieving predictive accuracy that surpasses the conventional visual methods. These results highlight the potential of neural network-based approaches to overcome limitations of traditional visual reconstruction, particularly in complex marine environments where occlusions from breaking waves or other disturbances are common. The findings also emphasize the importance of multi-scale spatial-temporal feature fusion in achieving robust and high-fidelity surface reconstruction under challenging observational conditions.

*2.5 3D internal velocity reconstruction*

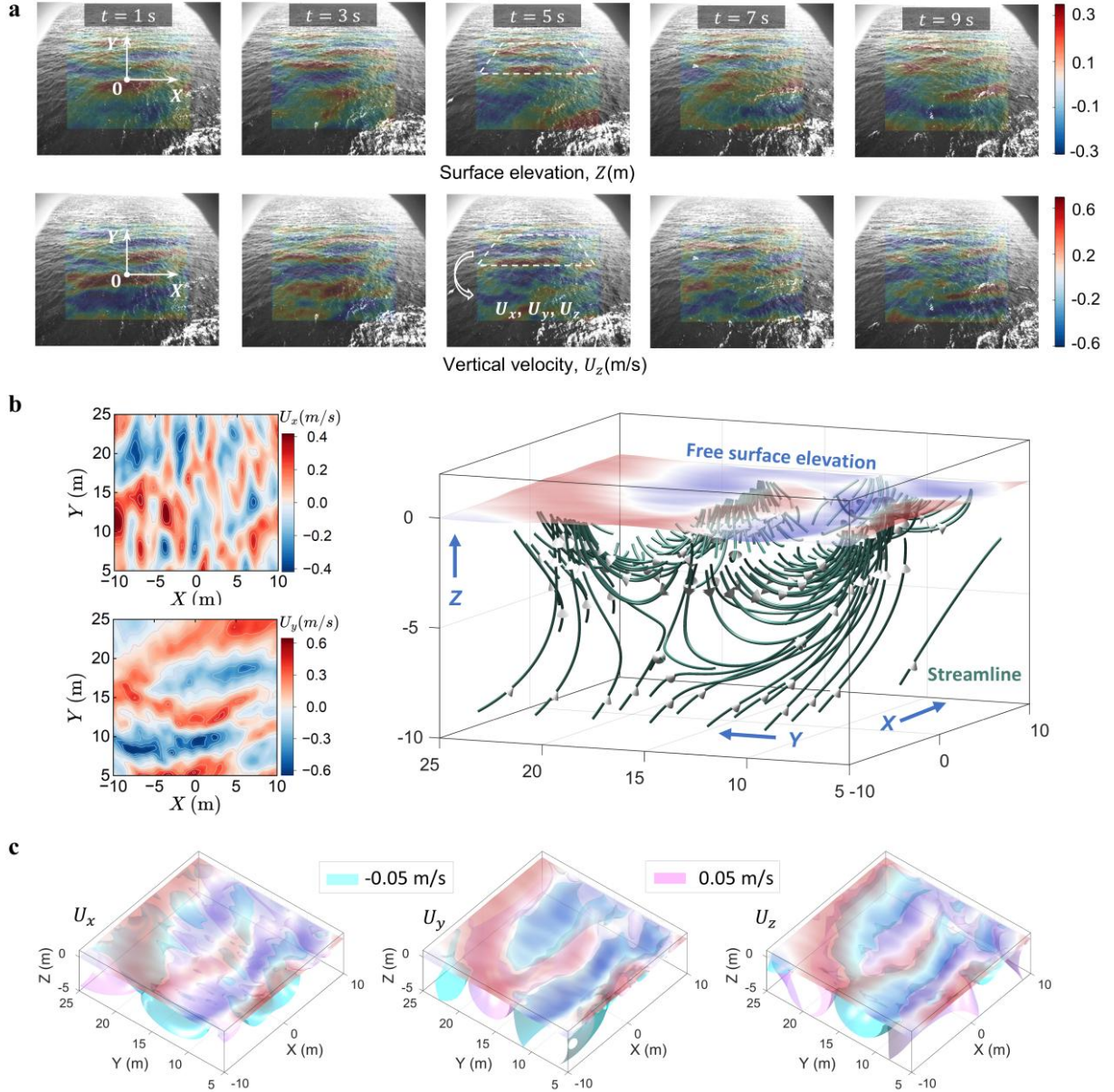

**Fig. 6 3D velocity field reconstruction. a** Time-sequential visualization of reconstructed free surface at different instants, with the corresponding vertical velocity distribution superimposed on the free surface under A1 condition. **b** 3D velocity fields extracted from $t = 5$ s, including horizontal velocity components $(U_x, U_y)$ at free surface and streamlines in wave field. **c** Isosurface of the 3D velocity field at $t = 5$ s.

After obtaining the spatiotemporal evolution data of the wave free surface, the vertical velocity, $U_z$, at the surface can be straightforwardly derived within the framework of linear wave theory. In contrast, the determination of horizontal velocity components, $\vec{U}_\eta = (U_x, U_y)$ at free surface, requires solving the nonlinear Zakharov equation[48], which considerably increases both the mathematical complexity. To address this, a simplified and effective nonlinear method has been developed to derive $\vec{U}_\eta$, as expressed by

$$U_z = \frac{1}{1 + |\nabla \eta|^2} \left( \frac{\partial \eta}{\partial t} + \nabla \eta \int_0^t \nabla \mathcal{T} \, dt \right) \quad (1)$$

$$\vec{U}_\eta = \int_0^t \nabla \mathcal{T}\, dt - \nabla\eta U_z \qquad (2)$$

in which $\mathcal{T}$ is given as

$$\mathcal{T} = -g\eta + \frac{1}{2}\frac{1}{1+|\nabla\eta|^2}\left(\frac{\partial\eta}{\partial t}\right)^2 \qquad (3)$$

where $g$ is the gravitational acceleration.

Based on the theoretical framework described above, we numerically reconstruct the time-resolved 3D velocity field $(U_x, U_y, U_z)$ from the sequentially recovered free surface elevation data $\eta(X,Y,t)$. Spatial gradients $\nabla\eta$ are computed using central differencing schemes, while temporal derivatives $\partial\eta/\partial t$ are estimated via a backward three-point finite difference method. The time integration of $\int_0^t \nabla\mathcal{T}\,dt$ is performed in the frequency domain using Fourier-based integration with regularization to mitigate low frequency singularities.

Fig. 6a provides the visualization of the reconstructed wave elevation dynamics and 3D velocity fields under condition A1. In the time-sequential snapshots, the wave elevation evolves from $t=1$ s to $t=9$ s, with the superimposed vertical velocity field $(U_z)$ highlighting the kinematic response of the free surface. The calculated vertical velocity reaches magnitudes up to ±0.6m/s, consistent with the expected scale of orbital motion induced by surface waves. The spatiotemporal evolution demonstrates coherent propagation of wave crests and troughs, while the associated vertical motions remain phase-locked with the surface elevation. The horizontal velocity components $(U_x, U_y)$ at $t=5$ s, along with the vertical velocity at the same time instant, are extracted to illustrate their spatial distribution and correlation (Fig. 6b). The horizontal velocity fields exhibit alternating banded structures aligned with the direction of wave propagation.

Once the 3D velocity field at the free surface is obtained, the subsurface velocity field can be reconstructed within the framework of potential flow theory. For viscous flows, such a reconstruction is generally ill-posed due to the lack of complete boundary and initial conditions, resulting in non-unique solutions. In contrast, potential flows do not suffer from this limitation, as a bottom boundary condition can be imposed to enforce an exponentially decaying velocity with depth. The reconstruction is carried out by constructing the fundamental solutions of the Laplace's equation for the velocity potential and ensuring that the free surface boundary conditions are satisfied, thereby yielding a subsurface velocity field consistent with the measured surface velocities. The velocity potential $\Phi(X,Y,Z,t)$ is expressed as a Fourier series in the horizontal directions,

$$\Phi(X,Y,Z,t) = \sum_{n=-N}^{N}\sum_{m=-M}^{M} a_{nm}(t) e^{\kappa_{nm} Z} e^{i(k_n X + k_m Y)} \qquad (4)$$

with coefficients $a_{nm}(t)$ modulated by the vertical decay factor $e^{\kappa_{nm} Z}$ and the horizontal wavenumbers $e^{i(k_n X + k_m Y)}$, where $\vec{\kappa}_{nm}$ has $\kappa_{nm} = (k_n^2 + k_m^2)^{1/2}$.

A fully connected neural network is constructed to map the input time $t$ to the coefficients $a_{nm}(t)$, as shown in Fig. 1. The network is trained by minimizing a physics-informed loss function that enforces consistency between the velocity reconstructed from $a_{nm}(t)$ and the observed velocity $\boldsymbol{U}$ at the free surface $z = \eta(X,Y,t)$, while penalizing high-frequency components through a Tikhonov regularization term weighted by $\lambda$.

$$\mathcal{L}_a(a_{nm}(t)) = \frac{1}{2}\left\|\nabla\Phi|_{z=\eta(X,Y,t)} - \boldsymbol{U}|_{z=\eta(X,Y,t)}\right\|^2 + \frac{1}{2}\lambda \sum_{n=-N}^{N}\sum_{m=-M}^{M}\|\kappa_{nm}a_{nm}\|^2 \quad (5)$$

After obtaining the optimal coefficients $a_{nm}(t) = \arg\min_{a_{nm}(t)} \mathcal{L}_a$, the internal velocity field is reconstructed by computing the spatial derivatives of the potential function defined in Eq. (4). The right side of Fig. 6b presents the streamlines of the 3D velocity field, providing a clear visualization of the particle trajectories within the flow. The streamlines show that the particles follow arc paths, highlighting the underlying transport mechanisms and the complex 3D wave motion. Fig. 6c shows the 3D isosurface of the instantaneous wave-induced velocity field. Regions of higher velocity appear beneath the wave crest and gradually weaken with depth. These methods reveal the organized spatial structure of the subsurface flow, providing direct insight into the internal hydrodynamics and energy distribution below the surface waves. Building on this, we achieve neural reconstruction of 3D wave hydrodynamics using ocean-wave video sequences. Even in irrotational flow, the subsurface motion dictates essential processes, including the exponential attenuation of orbital velocities with depth, the phase-resolved distribution of kinetic energy, and the spatial gradients of dynamic pressure induced by propagating waves. These insights offer an unprecedented view of the flow structures associated with surface waves, with implications for wave mechanics, coastal hydrodynamics, and ocean engineering applications.

## 3. Discussion

This study presents WHVS-Net, a neural framework for accurate 3D reconstruction of wave hydrodynamics, including both the free surface and velocity field. Unlike conventional vision-based methods that depend on dense stereo matching and are computationally intensive and sensitive to occlusions, WHVS-Net uses an attention-augmented pyramid architecture tailored to the multi-scale and temporally coherent characteristics of ocean waves. We employ implicit encoding of camera parameters to facilitate fast matching in the latent image space. Multi-scale attention captures fine-scale free surface structures, while spatiotemporal continuity enables the network to learn the dynamic motion patterns of ocean waves.

Experiments conducted under real-sea conditions demonstrate millimetre-level accuracy in wave height prediction and dominant frequency errors of less than 0.01 Hz. The model further captures fine-scale wave dynamics by accurately reproducing the high-frequency spectral power law of $f^{-5}$. Through the integration of spatial multi-scale features with temporal continuity, the framework achieves high-accuracy surface reconstruction even in the presence of visual occlusions. Leveraging a stereo vision dataset, the neural network consistently surpasses conventional vision-based methods and demonstrates strong generalization by reliably reconstructing occluded scenarios. In addition, a velocity field solver is developed for the free surface, and a potential-flow-based regularized least-squares formulation is applied to enforce consistency with velocity observations. This approach enables accurate and high-fidelity reconstruction of the 3D internal flow fields of ocean waves.

Furthermore, expanding the data acquisition framework to incorporate a broader spectrum of wave conditions, illumination scenarios, and geographical settings will facilitate the construction of a more diverse, representative, and benchmark-ready dataset ecosystem. Finally, the integration of multi-modal sensing strategies—combining visual, radar[49, 50] or LiDAR[51, 52]—may offer a promising path toward enhanced robustness and cross-scenario generalization. Collectively, these research directions are expected to advance neural network-based wave hydrodynamic reconstruction from experimental proof-of-concept to scalable, field-deployable systems, thereby making a substantive contribution to quantitative sea observation and ocean physics.

## 4. Methods

### 4.1 Feature extraction

Given a pair of input RGB images $I_L, I_R \in \mathbb{R}^{V \times U \times 3}$, an initial normalization step is applied to scale the intensity values to the range $[-1, 1]$. The normalized images, denoted as $I_{L,norm}$ and $I_{R,norm}$, are subsequently fed into a convolutional backbone to extract dense feature representations. Specifically, the left and right feature maps are extracted via a shared *ResNet*[45] encoder:

$$F_L, F_R = ResNet(I_{L,norm}, I_{L,norm}) \in \mathbb{R}^{H \times W \times D} \tag{6}$$

In parallel, multi-scale feature maps are computed from the left image using a hierarchical multi-scale ResNet-based encoder:

$$F'_s = MultiScaleResNet(I_{L,norm}) \in \mathbb{R}^{\frac{H}{2^s} \times \frac{W}{2^s} \times D}, s \in \{0,1,2,...\} \tag{7}$$

To incorporate camera extrinsic parameters (rotation $R$ and translation $T$), we introduce an extrinsic FiLM module. This module maps the extrinsic matrix into channel-wise affine parameters $(\gamma, \beta)$ via a multi-layer perceptron:

$$ExtrinsicFiLM(F, R, T) = F \odot (1 + \gamma) + \beta \tag{8}$$

where $(\gamma, \beta)$ are generated from the flattened extrinsic matrix $[vec(R), T] \in \mathbb{R}^{12}$. By extending camera parameters into a higher-dimensional space and fusing image features within a multi-scale low-dimensional embedding space, efficient feature matching and reconstruction are achieved. The resulting parameter modulation is applied to feature maps, facilitating network adaptation to varying camera poses.

*4.2 Multi-scale disparity refinement*

To compute the horizontal correlation between the left and right feature maps $F_L, F_R$ in the latent space, a row-wise latent normalized correlation volume $C_{1D}(i,j,k)$ is constructed. For each row index $i$, the similarity between pixel $j$ in the left feature map and pixel $k$ in the right feature map is defined as:

$$C_{1D}(i,j,k) = \frac{1}{\sqrt{D}} \sum_{d=1}^{D} F_L(i,j,d) \cdot F_R(i,k,d) \in \mathbb{R}^{H \times W \times W} \tag{9}$$

where $D$ denotes channel dimension of the feature map. This operation enforces a One-dimensional matching constraint by restricting the dot-product computation to feature vectors along the same image row. To ensure physically valid stereo correspondence, the matching is constrained to only consider candidates to the left of the current pixel in the left image, i.e.,

$$C_{1D}(i,j,k) = 0, if\ k > j \tag{10}$$

The similarity scores along each row are then normalized via a softmax function to obtain a matching probability distribution:

$$W_{disp}(i,j,k) = softmax(C_{1D}(i,j,k)) = \frac{e^{C_{1D}(i,j,k)}}{\sum_{k=1}^{W} e^{C_{1D}(i,j,k)}} \in \mathbb{R}^{H \times W \times W} \tag{11}$$

Let $P = [0,1,2,\ldots,W-1] \in \mathbb{R}^W$ represent the horizontal pixel indices in the right image. The sub-pixel disparity coordinate $(i, j)$ is then computed as the expected matching location:

$$Coor_R(i,j) = \sum_{k=1}^{W} W_{disp}(i,j,k) \cdot P_k \in \mathbb{R}^{H \times W} \tag{12}$$

Finally, the coarse disparity cues $d_{match}$ is given by:

$$d_{match}(i,j) = j - Coor_R(i,j) \in \mathbb{R}^{H \times W} \tag{13}$$

This formulation enables efficient and differentiable estimation of dense disparity through soft-attention over row-wise correlation volumes. Disparity maps are highly correlated with the source images in spatial structure. To refine the disparity predictions, an attention-guided[53] correction mechanism is introduced, wherein the reference feature map $F'_s$ provides scale-specific guidance to the coarse disparity cues $d_{match}$. This operation is implemented via a self-attention that adaptively reweights disparity values based on spatial consistency information:

$$d_{scale,s} = softmax\left(\frac{F'_s F'^T_s}{\sqrt{D}}\right) d_{match,s} \in \mathbb{R}^{\frac{H}{2^s} \times \frac{W}{2^s}}, s \in \{0,1,2,\ldots\} \tag{14}$$

Each disparity cues is replaced by a weighted sum over all spatial locations, where the attention weights are dynamically derived from the feature similarity across the reference map. This operation enables spatial consistency to inform local reconstruction, effectively suppressing irrelevant noise and enhancing the coherence and accuracy of the disparity field.

*4.3 Temporal fusion for video stereo*

To construct a disparity-difference consistency mask, given three consecutive initial disparity maps $d_{t-1}^0, d_t^0, d_{t+1}^0$, the mask between any two frames $l, k \in [t-1, t, t+1]$ is defined as:

$$m_{a \to b}(u, v) = exp\left(-\frac{|d_a^0(u,v) - d_b^0(u,v),|}{\sigma_d}\right) \quad (15)$$

where $\sigma_d$ controls the decay rate of the mask. $m \in [0, 1]$ represents the confidence of disparity consistency between adjacent frames at pixel.

For the current frame $d_t$ the disparity is refined by incorporating residual information from its neighboring frames:

$$d_t = d_t^0 + \tau_{-1} m_{t-1 \to t}(d_{t-1}^0 - d_t^0) + \tau_1 m_{t+1 \to t}(d_{t+1}^0 - d_t^0) \quad (16)$$

where $\tau_{-1}, \tau_1 \in [0, 1]$ denote the residual fusion weights for the previous and the next frame, respectively.

*4.4 Comparison with established models*

During network inference, the predicted disparity map $d_{pred}$ exhibits a distinct spatial pattern: along the vertical axis, regions with smaller $v$ values (i.e., waves closer to the camera) tend to have larger disparities and higher variance, whereas regions with larger $v$ values (i.e., more distant waves) show smaller disparities and reduced variance. This spatial heterogeneity suggests that disparity errors should be evaluated with position-dependent sensitivity.

To account for this, a position-aware weighted L1 disparity loss function is designed to prioritize accuracy in distant (low-disparity) regions while allowing greater tolerance for near-field (high-disparity) regions. The loss is formulated as:

$$L_{disp} = \frac{1}{|\Omega|} \sum_{(u,v) \in \Omega} \omega(v) \cdot |d_{pred}(u,v) - d_{GT}(u,v)| \quad (17)$$

where $\Omega$ denotes the set of valid pixels, and the vertical position-dependent weighting function $\omega(v)$ models the physical variation of disparity with respect to image row coordinate $v$:

$$\omega(v) = \alpha + (1 - \alpha)\left(1 - \frac{v}{V}\right)^\beta \quad (18)$$

where, $\alpha = 0.3$ defines the minimum weight assigned to pixels near the bottom of the image, and a quadratic decay is applied with $\beta = 2$.

In Table 2, we compare WHVS-Net with Raft-stereo[32], GMStereo[33], Selective-Stereo[34], MonSter[54] an DEFOM-Stereo[55] in terms of model parameters, inference time, and performance on the proposed wave benchmark for non-sequential inputs. Inference was performed on a single NVIDIA 4090 GPU at a resolution of 1632 × 1280. We evaluate performance using far-field and full-field endpoint errors, as well as Bad 1px and D1 metrics. The results indicate that WHVS-Net delivers the lowest errors and the fastest inference compared to the other methods, while maintaining a small parameter size.

Table 2 Quantitative evaluation on WASS dataset benchmark

| Model | EPE(pixel) | | Bad 1px(%) | D1(%) | Param (M) | Runtime (s) |
|---|---|---|---|---|---|---|
| | Full-field | Far-field | | | | |
| Raft-stereo[32] | 2.13 | 3.18 | 28.42 | 20.59 | 11.1 | 4.41 |
| GMStereo[33] | 2.24 | 3.97 | 30.36 | 21.32 | 7.4 | 1.78 |
| Selective-Stereo[34] | 1.99 | 2.35 | 27.14 | 19.86 | 11.7 | 2.19 |
| MonSter[54] | 2.09 | 4.02 | 29.31 | 20.18 | 53.4 | 3.40 |
| DEFOM-Stereo[55] | 1.86 | 3.71 | 26.64 | 19.38 | 47.3 | 2.45 |
| WHVS-Net | **1.72** | **2.32** | **25.81** | **16.41** | 7.9 | **1.35** |

In Table 3, ablation experiments are conducted to evaluate the importance and effectiveness of each key module in WHVS-Net. The configurations include using the full model, removing the temporal module, reducing the multi-scale depth, and eliminating the attention and reference mechanisms. Removing this module increases the temporal consistency metric $\sigma_t$, the temporal mean RMSE of disparity differences between consecutive frames, indicating the crucial role of temporal information in achieving stable reconstruction. Reducing multi-scale depth markedly degrades far-field accuracy, confirming their ability to capture wave features across scales. Eliminating attention leads to the worst performance, highlighting its critical role in feature aggregation. Together, these results verify the necessity of all modules and their complementary contributions to the robustness and accuracy of WHVS-Net.

Table 3 Ablation experiments of WHVS-Net

| Setup | EPE(pixel) | | Bad 1px(%) | D1(%) | $\sigma_t$(pixel) |
|---|---|---|---|---|---|
| | Full-field | Far-field | | | |
| Full | **1.65** | **2.29** | **24.12** | **16.01** | **0.32** |
| W/o temporal | 1.72 | 2.32 | 25.81 | 16.41 | 0.46 |
| MS depth 2 | 1.88 | 3.41 | 27.24 | 19.11 | — |
| MS depth 1 | 2.29 | 4.04 | 31.11 | 21.77 | — |
| W/o attn.+ref. | 2.77 | 5.15 | 36.97 | 27.51 | — |

*4.5 Derivation of Eqs. (1) and (2)*

Water wave problems are conventionally resolved in the framework of potential flow theory. It is assumed that (1) the fluid is inviscid, irrotational and incompressible and that (2) the effects of the surface tension, dissolved gases, cavitation, density, and temperature gradients of the water are negligible. The condition at the free surface, $\eta$, includes the dynamic free surface boundary condition,

$$\eta = -\frac{1}{g}\frac{d\Phi}{dt}\bigg|_{z=\eta} - \frac{1}{2g}\nabla\Phi|_{z=\eta} \cdot \nabla\Phi|_{z=\eta} - \frac{1}{2g}U_z^2 \qquad (19)$$

and the kinematic free surface boundary condition,

$$\frac{\partial \eta}{\partial t} + \nabla \eta \cdot \nabla \Phi|_{z=\eta} = U_z \tag{20}$$

where $g$ is the gravitational acceleration and where $U_z$ is the vertical velocity at the free surface. $\Phi$ is the potential function of water waves. For convenience of the subsequent derivation, the free surface velocity potential function defined by

$$\Phi^s(x,t) = \Phi(x, z = \eta, t) \tag{21}$$

By substituting it into Eqs. (19) and (20), the free surface boundary conditions become

$$\frac{\partial \Phi^s}{\partial t} = -g\eta - \frac{1}{2}|\nabla \Phi^s|^2 + \frac{1}{2}(1 + |\nabla \eta|^2)U_z^2 \tag{22}$$

$$\frac{\partial \eta}{\partial t} = (1 + |\nabla \eta|^2)U_z - \nabla \Phi^s \nabla \eta \tag{23}$$

Because $\eta$ can be obtained by visual reconstruction, $\Phi^s$ can be determined from

$$\frac{\partial \Phi^s}{\partial t} = -g\eta - \frac{1}{2}|\nabla \Phi^s|^2 + \frac{1}{2}(1 + |\nabla \eta|^2)\left(\frac{\partial \eta/\partial t + \nabla \Phi^s \nabla \eta}{(1 + |\nabla \eta|^2)}\right)^2 \tag{24}$$

In the above equation, an iterative procedure is required to obtain the solution of $\Phi^s$. However, since the term $|\nabla \Phi^s|^2$ exhibits relatively weak nonlinearity, it can be reasonably approximated and simplified as

$$\Phi^s \approx \int_0^t \left(-g\eta + \frac{1}{2}\frac{1}{1+|\nabla \eta|^2}\left(\frac{\partial \eta}{\partial t}\right)^2\right) dt = \int_0^t \mathcal{T}\, dt \tag{25}$$

Then, the nonlinear velocities at wave free surface are given as

$$U_z = \frac{1}{(1+|\nabla \eta|^2)}\left(\frac{\partial \eta}{\partial t} + \nabla \Phi^s \nabla \eta\right) = \frac{1}{1+|\nabla \eta|^2}\left(\frac{\partial \eta}{\partial t} + \nabla \eta \int_0^t \nabla \mathcal{T}\, dt\right) \tag{26}$$

$$\vec{U}_\eta = \nabla \Phi^s - \nabla \eta U_z = \int_0^t \nabla \mathcal{T}\, dt - \nabla \eta U_z \tag{27}$$

**Acknowledgment**

Financial supports from the National Natural Science Foundation of China (Grant No. 52478507, 523B2087, U24A20174) are greatly appreciated by the authors.